\documentclass[10pt,letterpaper]{article}
\usepackage[letterpaper,margin=1in]{geometry}
\usepackage[T1]{fontenc}
\usepackage[utf8]{inputenc}
\usepackage{amsmath,amsthm,mathtools}
\usepackage{amssymb}
\usepackage{newtxtext,newtxmath}
\usepackage{graphicx,booktabs,array,tabularx}
\usepackage[table,dvipsnames]{xcolor}
\usepackage{enumitem,microtype}
\usepackage[font=small,labelfont=bf]{caption}
\usepackage{url}
\usepackage[colorlinks=true,citecolor=blue,linkcolor=blue,urlcolor=blue]{hyperref}
\usepackage{bookmark}

\theoremstyle{remark}
\newtheorem{remark}{Remark}

\theoremstyle{definition}

\newcommand{\R}{\mathbb{R}}

\newcommand{\so}{\mathfrak{so}}

\DeclareMathOperator{\diag}{diag}

\newcommand{\Cl}{\mathcal{C}\ell}
\setlist{nosep,topsep=4pt}
\graphicspath{{figs/}}

\hypersetup{
    colorlinks=true,
    linkcolor=blue,
    citecolor=blue,
    urlcolor=blue
}
\title{\textbf{Subspace Clifford Networks}}
\author{
    Zhongping Ji \\
    \small \texttt{2026.10.27}
}
\date{}

\begin{document}

\maketitle

\begin{abstract}
Recent advances in Clifford Algebra Networks (CliffordNet) have demonstrated that formulating visual feature interactions through the Clifford Geometric Product ($uv = u \cdot v + u \wedge v$) endows vision backbones with rich algebraic expressivity, coupling scalar feature alignment with bivector geometry in a unified operator. In this work, we propose Subspace Clifford Networks, a reformulation that replaces global cyclic rolling with grouped subspace interactions. Instead of sampling sparse cyclic diagonals across the entire $D$-dimensional channel space, we project the features into an aligned space $\R^{\tilde{D}} = \bigoplus_{g=1}^{G} V_g$ partitioned into $G$ low-dimensional groups $V_g \cong \R^k$ with $k \in \{2, 3, 4\}$. Within each $k$-dimensional group, we compute all $\binom{k}{2}$ wedge coordinates and concatenate them with $\operatorname{SiLU}$-activated coordinate-wise products, while sandwiching the grouped interaction between learnable $1\times 1$ linear projections to enable cross-group channel mixing. This design reduces intermediate feature expansion from $2|\mathcal{S}|D$ (typically $10D$ for 5 shifts) down to $1.5\tilde{D}$ ($k=2$), $2.0\tilde{D}$ ($k=3$), or $2.5\tilde{D}$ ($k=4$). Furthermore, exploiting the small, fixed group size $k \in \{2, 3, 4\}$, we implement a fused CUDA kernel that evaluates the forward interaction and a recomputation-based analytical backward pass using thread-local variables without allocating global-memory buffers for elementwise intermediates. Finally, we extend the backbone from an isotropic columnar layout to a multi-resolution Hierarchical Clifford Network with configurable per-stage subspace dimensions. Experiments on CIFAR-100 show that the proposed models substantially reduce parameter count and training time across both regimes, while the hierarchical variants also improve classification accuracy over the evaluated shift-based hierarchical baselines.
\end{abstract}

\section{Introduction}

Modern computer vision backbones—ranging from Convolutional Neural Networks (CNNs) \cite{sandler2018mobilenetv2, ma2018shufflenet} to Vision Transformers (ViTs) \cite{dosovitskiy2021image, touvron2021training} and State-Space Models (SSMs) \cite{gu2023mamba, zhu2024vision, liu2024vmamba}—commonly follow the ``MetaFormer'' design pattern \cite{yu2022metaformer}, interleaving spatial token mixers with channel-mixing Feed-Forward Networks (FFNs). Recently, CliffordNet \cite{ji2026cliffordnet} explored an alternative design based on Geometric Algebra (Clifford Algebra) \cite{hestenes1966space, dorst2007geometric}. By modeling feature evolution through the Clifford Geometric Product ($uv = u \cdot v + u \wedge v$), CliffordNet combines symmetric feature alignment (via an inner-product branch) and anti-symmetric structural variation (via an exterior wedge-product branch), achieving competitive accuracy without separate FFN blocks.

Operationalizing exterior wedge interactions in a $D$-dimensional channel space ($\R^D$), however, presents a clear computational trade-off. While coordinate-wise products scale linearly with channel dimension $D$, evaluating all pairwise wedge components $u \wedge v$ spans $\binom{D}{2} = \frac{D(D-1)}{2}$ pairs, incurring $\mathcal{O}(D^2)$ intermediate channels. To keep the interaction width linear in $D$, the original CliffordNet \cite{ji2026cliffordnet} introduced a \textit{Sparse Rolling Interaction} mechanism, pairing each channel $c$ with a subset of cyclically shifted channels $(c+s) \bmod D$ for $s \in \mathcal{S}$ (e.g., $\mathcal{S}=\{1, 2, 4, 8, 16\}$) via cyclic tensor shifts (\texttt{torch.roll}).

In practice, this shift-based implementation has three main costs:
\begin{enumerate}
    \item \textbf{Sparse Pair Coverage:} Rolling samples only a predefined subset of channel pairs $(c, (c+s)\bmod D)$ across the global channel dimension, rather than evaluating all pairwise wedge interactions within local channel groups.
    \item \textbf{Channel Expansion and Projection Overhead:} For $|\mathcal{S}|=5$ shifts in full mode (inner + wedge), the rolling strategy concatenates $2|\mathcal{S}| = 10$ feature maps, expanding the intermediate channel dimension to $10D$. Consequently, the subsequent contraction projection $\mathcal{P}$ requires a $10D \times D$ linear layer, which increases parameter count and multiply-accumulate operations (MACs).
    \item \textbf{Memory-Bound Execution Latency:} In standard deep learning frameworks, executing multiple cyclic shifts, element-wise subtractions, products, and wide concatenations launches multiple separate GPU kernels. These operations allocate and read/write intermediate tensors in global device memory, increasing both latency and activation memory footprint during training.
\end{enumerate}

In this paper, we address these costs by proposing \textbf{Subspace Clifford Networks}. Instead of sampling sparse cyclic shifts across all $D$ channels, we partition the projected feature channels into $G = \tilde{D}/k$ low-dimensional groups of size $k \in \{2, 3, 4\}$. Within each $k$-dimensional group, we evaluate all $\binom{k}{2}$ pairwise wedge coordinates alongside $k$ $\operatorname{SiLU}$-activated coordinate-wise products. To enable cross-group channel communication, we place the grouped interaction between two learnable $1\times 1$ linear projections: an input projection mixes the $D$ input channels and generates the state and context streams prior to grouping, while an output projection maps the concatenated group outputs back to $\R^D$.

This grouped formulation substantially reduces the intermediate channel width. Even at $k=4$ (which produces $\binom{4}{2}=6$ wedge coordinates and $4$ inner coordinates per group), the intermediate channel dimension is $\frac{4+6}{4}\tilde{D} = 2.5\tilde{D}$—a \textbf{75\% reduction} compared to the $10D$ expansion of 5-shift rolling. Moreover, because the group size $k \in \{2, 3, 4\}$ is small and fixed, the element-wise operations can be fused into a single CUDA kernel. We implement a templated CUDA kernel that computes context differencing, within-group wedge and inner terms, $\operatorname{SiLU}$ activation, and a recomputation-based analytical backward pass using thread-local variables without allocating intermediate global-memory buffers. Finally, we extend the isotropic CliffordNet backbone to a multi-resolution \textbf{Hierarchical Clifford Network}, supporting spatial downsampling and stage-specific group sizes (e.g., $(k_1, k_2, k_3) = (2, 3, 4)$).

Our main contributions are summarized as follows:
\begin{itemize}
    \item \textbf{Grouped Subspace Clifford Interaction:} We reformulate high-dimensional Clifford interactions by decomposing the projected channel space into $G$ groups of dimension $k \in \{2, 3, 4\}$ coupled with input and output $1\times 1$ linear projections, providing complete within-group wedge pair coverage while reducing post-projection channel width by up to $75\%$ compared to 5-shift rolling.
    \item \textbf{Fused CUDA Kernel Implementation:} We develop a templated CUDA extension that fuses the forward and recomputation-based backward passes for $k \in \{2, 3, 4\}$, avoiding intermediate elementwise buffers in device memory and achieving a \textbf{17.2$\times$ operator speedup} and a \textbf{50\% reduction} in peak extra allocated memory for the evaluated $k=4$ benchmark.
    \item \textbf{Hierarchical \& Hybrid-Subspace Architecture:} Beyond the isotropic backbone, we introduce a multi-stage hierarchical CliffordNet architecture equipped with a contiguous $5\times 5$ \texttt{SpatialDiffusion} module, supporting configurable per-stage subspace dimensions across multi-resolution feature pyramids.
\end{itemize}

\section{Related Work}

\textbf{Clifford Algebra in Deep Learning.} Geometric Algebra (Clifford Algebra) \cite{hestenes1966space, dorst2007geometric} provides a mathematical framework encompassing complex numbers, quaternions, and exterior algebra. Early Clifford neural networks \cite{brandstetter2023clifford, brehmer2023geometric} primarily targeted 3D physics simulations and PDE modeling with $\mathrm{E}(3)$-equivariance, parameterizing network weights as full multivectors (e.g., $2^4=16$ components in $\Cl_{4,0}$). While expressive, multivector-valued weights increase matrix multiplication cost exponentially with algebra dimension. Recently, CliffordNet \cite{ji2026cliffordnet} applied a lightweight Clifford interaction over real-valued 2D vision feature maps, combining coordinate-wise inner products and shifted wedge differences without separate FFN blocks. Our work builds upon CliffordNet, replacing its cyclic shift approximation with grouped subspace interactions and a fused CUDA kernel.

\textbf{Efficient Vision Backbones \& Channel Mixing.} Modern lightweight backbones achieve efficiency through structural priors. MobileNetV2 \cite{sandler2018mobilenetv2} employs inverted residuals with depthwise separable convolutions, while ShuffleNetV2 \cite{ma2018shufflenet} introduces channel split and shuffle operations to facilitate cross-group information exchange. Vision Transformers \cite{dosovitskiy2021image, touvron2021training}, modernized hierarchical ConvNets such as ConvNeXt \cite{liu2022convnet}, and Vision SSMs \cite{zhu2024vision, liu2024vmamba} combine spatial token mixing with channel-mixing MLPs (typically using a $4\times$ channel expansion ratio). In contrast, our approach sandwiches low-dimensional ($k \in \{2, 3, 4\}$) subspace Clifford interactions—combining both coordinate-wise diagonal products and all $\binom{k}{2}$ antisymmetric off-diagonal wedge terms—between $1\times 1$ projections, maintaining a compact expansion ratio of $1.5\times$ to $2.5\times$.

\section{Methodology}

In this section, we first review the formulation and sparse rolling realization of the original CliffordNet \cite{ji2026cliffordnet} (Sections \ref{sec:review_framework} and \ref{sec:review_rolling}). We then present our \textbf{Subspace Clifford Interaction} (Section \ref{sec:subspace_gp}), the modified \textbf{CliffordAlgebraBlock} and \textbf{Hierarchical Architecture} (Section \ref{sec:architecture}), and our \textbf{Fused CUDA Kernel} (Section \ref{sec:cuda_fusion}).

\subsection{Recap: Feature Evolution and the Clifford Interaction Ansatz}
\label{sec:review_framework}

Let $\mathbf{I} \in \R^{C \times H_0 \times W_0}$ denote an input image. Rather than flattening image patches into a 1D sequence, CliffordNet preserves the 2D spatial grid $\Omega = \{1, \dots, H\} \times \{1, \dots, W\}$. The latent representation at layer $l$ is modeled as a $D$-dimensional vector field $\mathbf{H}^{(l)} : \Omega \to \R^D$, or in tensor form $\mathbf{X}^{(l)} \in \R^{D \times H \times W}$.

\textbf{Continuous Dynamical Evolution Motivation.} CliffordNet motivates feature updates across layers as a discrete approximation to a dynamical system governed by the interaction between the current feature state $\mathbf{H}(t)$ and a spatial context field $\mathcal{C}(\mathbf{H}(t))$:
$$
\frac{\partial \mathbf{H}}{\partial t} = \mathcal{F}\Big(\mathbf{H}, \, \mathcal{C}(\mathbf{H})\Big)
$$
where $\mathcal{C}(\cdot)$ aggregates local neighborhood information over $\Omega$ via depthwise convolutions, and $\mathcal{F}(\cdot, \cdot)$ denotes the interaction operator. Given a binary mode factor $\lambda \in \{0, 1\}$, the context vector field $\mathbf{C}$ is defined as:
$$
\mathbf{C} = \mathcal{C}_{\text{loc}}(\mathbf{H}) - \lambda \cdot \mathbf{H}
$$
where $\lambda = 1$ defines the \textit{Differential Mode} (which heuristics in \cite{ji2026cliffordnet} analogize to a discrete Laplacian difference $\mathbf{C} \approx \Delta \mathbf{H}$) and $\lambda = 0$ defines the \textit{Absolute Mode} ($\mathbf{C} = \mathcal{C}_{\text{loc}}(\mathbf{H})$).

\textbf{The Clifford Interaction Ansatz.} Standard dot-product layers compress two vectors $\mathbf{u}, \mathbf{v} \in \R^D$ into a symmetric scalar $\mathbf{u} \cdot \mathbf{v}$. CliffordNet instead motivates $\mathcal{F}$ from the Clifford Geometric Product in $\Cl_{D,0}(\R)$:
$$
\mathbf{u}\mathbf{v} = \underbrace{\mathbf{u} \cdot \mathbf{v}}_{\text{Symmetric (Grade 0)}} + \underbrace{\mathbf{u} \wedge \mathbf{v}}_{\text{Anti-Symmetric (Grade 2)}}
$$
Given an orthonormal basis $\{\mathbf{e}_1, \dots, \mathbf{e}_D\}$ of $\R^D$, the exterior wedge product $\mathbf{u} \wedge \mathbf{v}$ is a bivector in $\Lambda^2(\R^D)$:
$$
\mathbf{u} \wedge \mathbf{v} = \sum_{1 \le i < j \le D} (u_i v_j - v_i u_j)(\mathbf{e}_i \wedge \mathbf{e}_j)
$$
Each coefficient $u_i v_j - v_i u_j$ represents the signed area of the parallelogram spanned by projections onto the $(\mathbf{e}_i, \mathbf{e}_j)$ plane. In neural network form, CliffordNet replaces the scalar inner product with a coordinate-wise product branch $\mathcal{D}(\mathbf{H}, \mathbf{C})$, concatenates it ($\oplus$) with a wedge branch $\mathcal{W}(\mathbf{H}, \mathbf{C})$, and applies a learnable $1\times 1$ contraction projection $\mathcal{P}$:
$$
\frac{\partial \mathbf{H}}{\partial t} = \mathcal{P}\Big( \mathcal{D}(\mathbf{H}, \mathbf{C}) \oplus \mathcal{W}(\mathbf{H}, \mathbf{C}) \Big)
$$

\subsection{Recap: Sparse Rolling Realization and Its Costs}
\label{sec:review_rolling}

Directly evaluating all $\binom{D}{2} = \frac{D(D-1)}{2}$ wedge coefficients in $\mathcal{W}(\mathbf{H}, \mathbf{C})$ produces $\mathcal{O}(D^2)$ channels. To keep the intermediate channel width linear in $D$, the original CliffordNet \cite{ji2026cliffordnet} samples a subset of channel pairs using a cyclic shift operator $\mathcal{T}_s : \R^D \to \R^D$, defined by $[\mathcal{T}_s(\mathbf{u})]_c = u_{(c+s) \bmod D}$.

For a predefined set of shift offsets $\mathcal{S} \subset \{1, \dots, D-1\}$ (e.g., $\mathcal{S} = \{1, 2, 4, 8, 16\}$), the shifted inner and wedge slices are computed via element-wise Hadamard products $\odot$:
$$
\mathcal{D}_s(\mathbf{H}, \mathbf{C}) = \operatorname{SiLU}\Big(\mathbf{H} \odot \mathcal{T}_s(\mathbf{C})\Big) \in \R^D
$$
$$
\mathcal{W}_s(\mathbf{H}, \mathbf{C}) = \mathbf{H} \odot \mathcal{T}_s(\mathbf{C}) - \mathcal{T}_s(\mathbf{H}) \odot \mathbf{C} \in \R^D
$$
Concatenating these slices across all $s \in \mathcal{S}$ yields $\mathbf{G}_{\text{raw}} \in \R^{2|\mathcal{S}|D}$, which is projected back to $\R^D$ via a $1\times 1$ convolution $\mathbf{G}_{\text{feat}} = \mathcal{P}(\mathbf{G}_{\text{raw}})$ and added to the residual stream via a Gated Geometric Residual (GGR) update:
$$
\mathbf{H}^{(l)} = \mathbf{H}^{(l-1)} + \boldsymbol{\gamma} \odot \Big( \operatorname{SiLU}(\mathbf{H}_{\text{ln}}) + \sigma\big(\mathbf{W}_{\text{gate}}[\mathbf{H}_{\text{ln}} \oplus \mathbf{G}_{\text{feat}}]\big) \odot \mathbf{G}_{\text{feat}} \Big)
$$

\textbf{Limitations of the Rolling Realization.}
\begin{remark}[Sparse Pair Coverage]
Let $\mathbf{M} = \mathbf{u}\mathbf{v}^\top - \mathbf{v}\mathbf{u}^\top \in \so(D)$ be the full skew-symmetric matrix of wedge coordinates. The rolling operator $\mathcal{T}_s$ samples $|\mathcal{S}|$ cyclic off-diagonals of $\mathbf{M}$. Rather than evaluating all pairs within local channel groups, rolling only evaluates pairs separated by offsets $s \in \mathcal{S}$.
\end{remark}
\begin{remark}[Channel Expansion and Memory Traffic]
When $|\mathcal{S}|=5$ (as in CliffordNet-Lite), the concatenated tensor $\mathbf{G}_{\text{raw}}$ has channel width $2|\mathcal{S}|D = 10D$. Consequently, the contraction projection $\mathcal{P} \in \R^{D \times 10D}$ requires $10D^2$ MACs per pixel, and evaluating the shifts in standard frameworks requires multiple tensor shift and concatenation operations in device memory.
\end{remark}

\subsection{Proposed Method: Subspace Clifford Interaction}
\label{sec:subspace_gp}

To reduce intermediate channel expansion while evaluating all pairwise wedge coordinates within local groups, we replace global cyclic shifts with a grouped \textbf{Subspace Clifford Interaction}.

\subsubsection{Channel Grouping and Pre-Projection Mixing}

Let $\mathbf{X}_{\text{sp}} \in \R^{D \times H \times W}$ be the spatially diffused feature map at a given layer, and let $k \in \{2, 3, 4\}$ denote the chosen group size. To ensure that the internal channel width is divisible by $k$ (for instance, when $D=32$ and $k=3$), we define the aligned channel dimension $\tilde{D}$ and the number of groups $G$ as:
$$
\tilde{D} = \left\lceil \frac{D}{k} \right\rceil \cdot k, \qquad G = \frac{\tilde{D}}{k}
$$
To mix channels across groups prior to partitioning, we apply Batch Normalization ($\operatorname{BN}$) followed by a learnable $1\times 1$ input projection $\mathbf{W}_{\text{in}} \in \R^{2\tilde{D} \times D}$ that generates both the state stream $\mathbf{Z}_1 \in \R^{\tilde{D} \times H \times W}$ and the initial context stream $\mathbf{Z}_2' \in \R^{\tilde{D} \times H \times W}$:
$$
[\mathbf{Z}_1, \, \mathbf{Z}_2'] = \mathbf{W}_{\text{in}} \operatorname{BN}(\mathbf{X}_{\text{sp}})
$$
We then compute the spatial context stream $\mathbf{Z}_2$ via a $3\times 3$ depthwise convolution and form the interaction streams $\mathbf{U}$ and $\mathbf{V}$ with $\lambda \in \{0, 1\}$:
$$
\mathbf{Z}_2 = \operatorname{DWConv}_{3\times 3}(\mathbf{Z}_2'), \qquad \mathbf{V} = \mathbf{Z}_2 - \lambda \mathbf{Z}_1, \qquad \mathbf{U} = \mathbf{Z}_1
$$
The spatial context branch $\operatorname{DWConv}_{3\times 3}$ allows the two streams $\mathbf{Z}_1$ and $\mathbf{Z}_2$ to differ across spatial neighborhoods, enabling nonzero wedge responses whenever the $k$-dimensional group vectors are not collinear. Note that by the bilinearity and anti-symmetry of the exterior product ($\mathbf{u}^{(g)} \wedge \mathbf{u}^{(g)} = \mathbf{0}$), the wedge branch satisfies:
$$
\mathbf{u}^{(g)} \wedge \big(\mathbf{z}_2^{(g)} - \lambda \mathbf{u}^{(g)}\big) = \mathbf{u}^{(g)} \wedge \mathbf{z}_2^{(g)}
$$
Thus, in exact real arithmetic, the differential parameter $\lambda$ only alters the coordinate-wise inner branch, while leaving the wedge branch invariant.

\subsubsection{Within-Group Interaction for $k \in \{2, 3, 4\}$}

At each spatial location $(h, w)$, we partition the aligned channel space $\R^{\tilde{D}}$ into $G$ contiguous $k$-dimensional coordinate subspaces:
$$
\R^{\tilde{D}} = \bigoplus_{g=1}^{G} V_g, \qquad V_g \cong \R^k
$$
where $\mathbf{u}^{(g)} = [u_{g,0}, \dots, u_{g,k-1}]^\top \in \R^k$ and $\mathbf{v}^{(g)} = [v_{g,0}, \dots, v_{g,k-1}]^\top \in \R^k$ denote the $g$-th group slices of $\mathbf{U}$ and $\mathbf{V}$. Within each group $V_g$, we compute two feature branches:

\textbf{1. Coordinate-Wise Inner Branch ($\mathbf{p}_{\text{inner}}^{(g)}$):}
Rather than summing the coordinate products into a single scalar $\sum_{i=0}^{k-1} u_{g,i} v_{g,i}$, we retain all $k$ coordinate-wise products and apply a $\operatorname{SiLU}$ activation:
$$
\mathbf{p}_{\text{inner}}^{(g)} = \operatorname{SiLU}\Big(\mathbf{u}^{(g)} \odot \mathbf{v}^{(g)}\Big) = \begin{bmatrix} \operatorname{SiLU}(u_{g,0} v_{g,0}) \\ \vdots \\ \operatorname{SiLU}(u_{g,k-1} v_{g,k-1}) \end{bmatrix} \in \R^k
$$

\textbf{2. Complete Within-Group Wedge Branch ($\mathbf{p}_{\text{wedge}}^{(g)}$):}
Identifying $V_g \cong \R^k$ with the vector part of $\Cl_{k,0}(\R)$ with standard basis $\{\mathbf{e}_0, \dots, \mathbf{e}_{k-1}\}$, the exterior product $\mathbf{u}^{(g)} \wedge \mathbf{v}^{(g)} \in \Lambda^2(\R^k)$ has $\binom{k}{2} = \frac{k(k-1)}{2}$ basis bivector coordinates:
$$
\mathbf{u}^{(g)} \wedge \mathbf{v}^{(g)} = \sum_{0 \le i < j < k} \underbrace{(u_{g,i} v_{g,j} - u_{g,j} v_{g,i})}_{w_{g,ij}} (\mathbf{e}_i \wedge \mathbf{e}_j)
$$
Evaluating all $\binom{k}{2}$ pairs within the group yields $\mathbf{p}_{\text{wedge}}^{(g)} \in \R^{\binom{k}{2}}$. Specifically, for $k \in \{2, 3, 4\}$, our implementation computes:

\begin{itemize}
    \item \textbf{2D Groups ($k=2$):} Each group has $\binom{2}{2}=1$ wedge coordinate:
    $$
    \mathbf{p}_{\text{wedge}}^{(g)} = \big[ u_{g,0}v_{g,1} - u_{g,1}v_{g,0} \big] \in \R^1
    $$
    Concatenated with $\mathbf{p}_{\text{inner}}^{(g)} \in \R^2$, each 2D group outputs $2+1=3$ channels, giving $C_{\text{out}} = \frac{3}{2}\tilde{D} = 1.5\tilde{D}$.

    \item \textbf{3D Groups ($k=3$):} Each group has $\binom{3}{2}=3$ wedge coordinates. Following the cyclic 3D cross-product convention $\mathbf{u}^{(g)} \times \mathbf{v}^{(g)}$ (with indices $(0,1), (1,2), (2,0)$, where $w_{g,20} = -w_{g,02}$), we compute:
    $$
    \mathbf{p}_{\text{wedge}}^{(g)} = \begin{bmatrix} u_{g,0}v_{g,1} - u_{g,1}v_{g,0} \\ u_{g,1}v_{g,2} - u_{g,2}v_{g,1} \\ u_{g,2}v_{g,0} - u_{g,0}v_{g,2} \end{bmatrix} \in \R^3
    $$
    Concatenated with $\mathbf{p}_{\text{inner}}^{(g)} \in \R^3$, each 3D group outputs $3+3=6$ channels, giving $C_{\text{out}} = \frac{6}{3}\tilde{D} = 2.0\tilde{D}$.

    \item \textbf{4D Groups ($k=4$):} Each group has $\binom{4}{2}=6$ wedge coordinates ordered lexicographically by $(i, j)$ for $0 \le i < j < 4$:
    $$
    \mathbf{p}_{\text{wedge}}^{(g)} = \begin{bmatrix} 
    u_{g,0}v_{g,1} - u_{g,1}v_{g,0} \\ 
    u_{g,0}v_{g,2} - u_{g,2}v_{g,0} \\ 
    u_{g,0}v_{g,3} - u_{g,3}v_{g,0} \\ 
    u_{g,1}v_{g,2} - u_{g,2}v_{g,1} \\ 
    u_{g,1}v_{g,3} - u_{g,3}v_{g,1} \\ 
    u_{g,2}v_{g,3} - u_{g,3}v_{g,2} 
    \end{bmatrix} \in \R^6
    $$
    Concatenated with $\mathbf{p}_{\text{inner}}^{(g)} \in \R^4$, each 4D group outputs $4+6=10$ channels, giving $C_{\text{out}} = \frac{10}{4}\tilde{D} = 2.5\tilde{D}$.
\end{itemize}

Concatenating the wedge and inner branches within each group and stacking all $G$ groups produces the output tensor $\mathbf{F}_{\text{gp}} = \operatorname{SubspaceGP}_k(\mathbf{Z}_1, \mathbf{Z}_2)$:
$$
\mathbf{F}_{\text{gp}} = \bigoplus_{g=1}^{G} \left( \mathbf{p}_{\text{wedge}}^{(g)} \oplus \mathbf{p}_{\text{inner}}^{(g)} \right) \in \R^{C_{\text{out}} \times H \times W}, \qquad C_{\text{out}} = G \left(k + \binom{k}{2}\right) = \frac{k+1}{2}\tilde{D}
$$

\subsubsection{Tensor Interpretation of the Group Representation}
\label{sec:tensor_interpretation}

To understand why we retain the $k$-dimensional vector $\mathbf{p}_{\text{inner}}^{(g)} = \operatorname{SiLU}(\mathbf{u}^{(g)} \odot \mathbf{v}^{(g)}) \in \R^k$ instead of collapsing it into a single scalar $\sum_{i=0}^{k-1} u_{g,i} v_{g,i} \in \R^1$, we can examine the second-order outer (dyadic) product $\mathbf{T}^{(g)} = \mathbf{u}^{(g)} \otimes \mathbf{v}^{(g)} = \mathbf{u}^{(g)}(\mathbf{v}^{(g)})^\top \in \R^{k \times k}$. Any such $k \times k$ matrix decomposes into a symmetric part $\mathbf{S}^{(g)}$ and a skew-symmetric part $\mathbf{A}^{(g)}$:
$$
\mathbf{T}^{(g)} = \underbrace{\frac{1}{2}\Big(\mathbf{u}^{(g)}(\mathbf{v}^{(g)})^\top + \mathbf{v}^{(g)}(\mathbf{u}^{(g)})^\top\Big)}_{\text{Symmetric Part } \mathbf{S}^{(g)}} + \underbrace{\frac{1}{2}\Big(\mathbf{u}^{(g)}(\mathbf{v}^{(g)})^\top - \mathbf{v}^{(g)}(\mathbf{u}^{(g)})^\top\Big)}_{\text{Skew-Symmetric Part } \mathbf{A}^{(g)} \cong \mathbf{u}^{(g)} \wedge \mathbf{v}^{(g)}}
$$
Because $\mathbf{A}_{ii}^{(g)} = 0$ on the main diagonal, the $k^2$ components of $\mathbf{T}^{(g)}$ partition into three subsets:
\begin{enumerate}
    \item \textbf{Main Diagonal Components ($k$ terms):} $\diag(\mathbf{T}^{(g)}) = \mathbf{u}^{(g)} \odot \mathbf{v}^{(g)} = [u_{g,0}v_{g,0}, \dots, u_{g,k-1}v_{g,k-1}]^\top \in \R^k$, which lie entirely in $\mathbf{S}^{(g)}$.
    \item \textbf{Antisymmetric Off-Diagonal Components ($\binom{k}{2}$ terms):} $w_{g,ij} = u_{g,i}v_{g,j} - u_{g,j}v_{g,i}$ ($i < j$), which form the basis coordinates of $\mathbf{A}^{(g)}$.
    \item \textbf{Symmetric Off-Diagonal Components ($\binom{k}{2}$ terms):} $s_{g,ij} = u_{g,i}v_{g,j} + u_{g,j}v_{g,i}$ ($i < j$).
\end{enumerate}

Our representation retains the $k$ diagonal products (with $\operatorname{SiLU}$ activation) and all $\binom{k}{2}$ antisymmetric off-diagonal components, while omitting the symmetric off-diagonal components. This produces
$$
d_k = k + \binom{k}{2} = \frac{k(k+1)}{2}
$$
output channels per group. Compared to a scalar dot product $\operatorname{Tr}(\mathbf{T}^{(g)}) = \sum_{i=0}^{k-1} u_{g,i}v_{g,i}$ that sums the diagonal into a single number, keeping the $k$ coordinate-wise products allows the subsequent $1 \times 1$ linear projection $\mathbf{W}_{\text{out}}$ to assign independent weights $w_i \in \R$ to each coordinate. Before applying $\operatorname{SiLU}$, a weighted sum $\sum_{i=0}^{k-1} w_i u_{g,i} v_{g,i} = (\mathbf{u}^{(g)})^\top \diag(\mathbf{w}) \mathbf{v}^{(g)}$ defines a diagonal bilinear form (which includes both positive and negative weights as in pseudo-Euclidean inner products $\Cl_{p,q}(\R)$). In our implemented module, $\operatorname{SiLU}$ introduces a coordinate-wise nonlinearity prior to $\mathbf{W}_{\text{out}}$, so the resulting projection acts as a learned linear combination of nonlinearly gated diagonal products and exact within-group wedge coordinates.

\subsection{Block Design and Hierarchical Multi-Resolution Architecture}
\label{sec:architecture}

\textbf{Reaction--Diffusion Motivation and Operator-Splitting Block Design.} 
From a continuous dynamical perspective, feature evolution across both spatial neighborhoods and feature channels can be viewed as a coupled \textit{reaction--diffusion system}:
$$
\frac{\partial \mathbf{X}}{\partial t} = \mathcal{D}_{\text{sp}}(\mathbf{X}) + \mathcal{R}_{\text{cli}}(\mathbf{X})
$$
where $\mathcal{D}_{\text{sp}}(\cdot)$ denotes a channel-wise spatial diffusion operator that propagates information across the 2D grid $\Omega$, and $\mathcal{R}_{\text{cli}}(\cdot)$ denotes a nonlinear cross-channel reaction operator driven by our subspace Clifford interaction. Rather than evaluating both terms simultaneously on the same input state, we discretize this system via a two-stage \textit{Lie--Trotter operator-splitting} scheme ($\mathbf{X} \xrightarrow{\mathcal{D}_{\text{sp}}} \mathbf{X}_{\text{sp}} \xrightarrow{\mathcal{R}_{\text{cli}}} \mathbf{Y}$), yielding a streamlined \texttt{CliffordAlgebraBlock} with two sequential residual sub-modules:
\begin{enumerate}
    \item \textbf{Spatial Diffusion Sub-step ($\mathcal{D}_{\text{sp}}$):} First, an optional \texttt{SpatialDiffusion} module aggregates spatial neighborhood context independently for each channel using a single $5\times 5$ depthwise convolution ($\operatorname{DWConv}_{5\times 5}$), followed by Batch Normalization ($\operatorname{BN}$) and in-place $\operatorname{SiLU}$ activation, scaled by a learnable vector $\boldsymbol{\gamma}_{\text{sp}} \in \R^{D \times 1 \times 1}$:
    $$
    \mathbf{X}_{\text{sp}} = \mathbf{X} + \boldsymbol{\gamma}_{\text{sp}} \odot \operatorname{SiLU}\Big( \operatorname{BN}\big(\operatorname{DWConv}_{5\times 5}(\mathbf{X})\big) \Big)
    $$
    When this explicit diffusion sub-step is omitted (\textit{SpatialDiffusion-free}), $\mathbf{X}_{\text{sp}} = \mathbf{X}$, and spatial propagation relies entirely on the $3\times 3$ context convolution within the subsequent reaction step.
    \item \textbf{Subspace Clifford Reaction Sub-step ($\mathcal{R}_{\text{cli}}$):} Taking the diffused state $\mathbf{X}_{\text{sp}}$ as input, the pre-projection $\mathbf{W}_{\text{in}}$ and context convolution $\operatorname{DWConv}_{3\times 3}$ generate $\mathbf{Z}_1, \mathbf{Z}_2$ to evaluate the nonlinear subspace interaction $\mathbf{F}_{\text{gp}} = \operatorname{SubspaceGP}_k(\mathbf{Z}_1, \mathbf{Z}_2) \in \R^{C_{\text{out}} \times H \times W}$. A learnable $1\times 1$ output projection $\mathbf{W}_{\text{out}} \in \R^{D \times C_{\text{out}}}$ then mixes the multivector reaction products back into $\R^D$, followed by LayerScale $\boldsymbol{\gamma}_{\text{blk}} \in \R^{D \times 1 \times 1}$ and Stochastic Depth ($\operatorname{DropPath}$):
    $$
    \mathbf{Y} = \mathbf{X}_{\text{sp}} + \operatorname{DropPath}\Big( \boldsymbol{\gamma}_{\text{blk}} \odot (\mathbf{W}_{\text{out}} \mathbf{F}_{\text{gp}}) \Big)
    $$
\end{enumerate}

\textbf{Complexity Analysis.} For a fixed group size $k \in \{2, 3, 4\}$, the subspace interaction $\operatorname{SubspaceGP}_k$ itself has $\mathcal{O}(D)$ linear complexity in the channel dimension, producing $C_{\text{out}} = \frac{k+1}{2}\tilde{D}$ channels. Including the dense $1\times 1$ input projection $\mathbf{W}_{\text{in}} \in \R^{2\tilde{D} \times D}$ and output projection $\mathbf{W}_{\text{out}} \in \R^{D \times C_{\text{out}}}$, the total pointwise projection cost per spatial token is:
$$
\text{Projection MACs} = 2D\tilde{D} + \frac{k+1}{2}D\tilde{D} = \frac{k+5}{2}D\tilde{D}
$$
which has quadratic channel complexity $\mathcal{O}(D^2)$, but reduces the post-projection cost from $10D^2$ (in 5-shift rolling) to $1.5D\tilde{D}$--$2.5D\tilde{D}$ (Table~\ref{tab:complexity}).

\begin{table}[th]
\centering
\caption{\textbf{Per-Token Interaction Complexity Comparison} between the ungrouped Full Diagonal + Wedge representation, Shift-based Rolling CliffordNet ($|\mathcal{S}|=5$ shifts), and our Subspace Clifford Interaction ($k \in \{2, 3, 4\}$) in full mode. Post-Proj MACs counts only the contraction projection $\mathcal{P}$ ($\mathbf{W}_{\text{out}}$).}
\label{tab:complexity}
\small
\begin{tabular}{lcccc}
\toprule
\textbf{Interaction Mechanism} & \textbf{Pair Coverage} & \textbf{Output Dim $C_{\text{out}}$} & \textbf{Post-Proj MACs} & \textbf{\texttt{roll} Shifts} \\
\midrule
Full Diagonal + Wedge (Ungrouped) & All $\binom{D}{2}$ Global Pairs & $D + \frac{D(D-1)}{2}$ & $\mathcal{O}(D^3)$ & $0$ \\
Sparse Rolling ($|\mathcal{S}|=5$) \cite{ji2026cliffordnet} & $|\mathcal{S}|D$ Cyclic Pairs & $10.0\,D$ & $10.0\,D^2$ & $10$ \\
\midrule
\rowcolor{gray!10} \textbf{2D Subspace ($k=2$)} & All Within-Group Pairs ($\binom{2}{2}=1$) & $\mathbf{1.5\,\tilde{D}}$ & $\mathbf{1.5\,D\tilde{D}}$ & $\mathbf{0}$ \\
\rowcolor{gray!10} \textbf{3D Subspace ($k=3$)} & All Within-Group Pairs ($\binom{3}{2}=3$) & $\mathbf{2.0\,\tilde{D}}$ & $\mathbf{2.0\,D\tilde{D}}$ & $\mathbf{0}$ \\
\rowcolor{gray!10} \textbf{4D Subspace ($k=4$)} & All Within-Group Pairs ($\binom{4}{2}=6$) & $\mathbf{2.5\,\tilde{D}}$ & $\mathbf{2.5\,D\tilde{D}}$ & $\mathbf{0}$ \\
\bottomrule
\end{tabular}
\end{table}

\textbf{From Isotropic to Hierarchical Clifford Networks.} While the original CliffordNet operates on an isotropic (single-resolution) grid, multi-scale feature hierarchies are beneficial for visual recognition. We therefore construct a multi-stage hierarchical \textbf{CliffordNet} comprising $S$ stages. Given an initial stem embedding, each stage $s \in \{1, \dots, S\}$ operates at spatial resolution $\frac{H_0}{P \cdot 2^{s-1}} \times \frac{W_0}{P \cdot 2^{s-1}}$ with configurable stage channel width $D_s$ (e.g., $(D_1, D_2, D_3) = (96, 128, 160)$) and depth $L_s$. Between consecutive stages, a \texttt{StageDownsample} module (a stride-2 $3\times 3$ depthwise convolution followed by BN, SiLU, and a $1\times 1$ projection) halves the spatial resolution and projects the channels from $D_s$ to $D_{s+1}$.

\textbf{Stage-Adaptive Hybrid Subspaces.} Because the group size $k$ is local to each block, it can vary across stages: $\mathbf{k} = (k_1, k_2, \dots, k_S)$ with $k_s \in \{2, 3, 4\}$. For example, early high-resolution stages can use $k_1 = 2$ ($1.5\tilde{D}_1$ intermediate channels) to reduce memory traffic, while deeper low-resolution stages can use $k_s = 3$ or $4$ ($2.0\tilde{D}_s$ or $2.5\tilde{D}_s$) to evaluate more wedge pairs per group.

\subsection{Fused CUDA Kernel Optimization}
\label{sec:cuda_fusion}

As summarized in Table~\ref{tab:complexity}, our grouped formulation eliminates \texttt{torch.roll} shifts and reduces $C_{\text{out}}$ from $10D$ to $1.5\tilde{D}$--$2.5\tilde{D}$. However, in eager-mode PyTorch, reshaping tensors into $(B, G, k, H, W)$ and executing separate elementwise subtractions, products, $\operatorname{SiLU}$ activations, and concatenations still launches multiple fine-grained GPU kernels and materializes intermediate tensors in global device memory. IO-aware implementations such as FlashAttention \cite{dao2022flashattention} demonstrate that reducing intermediate memory traffic and recomputing selected quantities during backpropagation can substantially improve both runtime and memory efficiency.

Motivated by this principle, and exploiting the small fixed group size $k \in \{2, 3, 4\}$, we implement a custom \textbf{fused CUDA kernel} parameterized via C++ compile-time templates $\langle K, \text{CLI\_MODE}, \text{CTX\_DIFF}\rangle$:
\begin{enumerate}
    \item \textbf{One-Thread-Per-Group Mapping:} We launch $B \times G \times H \times W$ parallel GPU threads. Each thread $idx$ is assigned to a single spatial location $(h, w)$ of a single $k$-dimensional group $(b, g)$.
    \item \textbf{Fused Forward Evaluation:} Using compile-time loop unrolling (`\#pragma unroll`), each thread loads $k$ floats from $\mathbf{Z}_1$ and $k$ floats from $\mathbf{Z}_2$ into thread-local arrays `float u[K]` and `float v[K]`. Context differencing ($v_i = z_{2,i} - \lambda z_{1,i}$), all $\binom{k}{2}$ wedge differences (which can leverage fused multiply-add instructions), and $k$ inner $\operatorname{SiLU}$ activations are evaluated directly on these thread-local variables and written once to the output tensor $\mathbf{F}_{\text{gp}}$ in device memory.
    \item \textbf{Recomputation-Based Backward Pass:} Following the IO-aware recomputation strategy \cite{dao2022flashattention}, rather than saving elementwise intermediate buffers (such as $\mathbf{V} = \mathbf{Z}_2 - \lambda\mathbf{Z}_1$ or pre-activation products $u_{g,i}v_{g,i}$) from the forward pass to global memory, the backward kernel reloads only $\mathbf{Z}_1, \mathbf{Z}_2$ and $\nabla_{\mathbf{F}_{\text{gp}}}\mathcal{L}$ and recomputes the local products on-the-fly in thread-local variables. Because arithmetic recomputation within a $k$-dimensional group is orders of magnitude faster than reading intermediate tensors from device memory, this strategy simultaneously reduces both peak activation memory and backward execution time. 
\end{enumerate}

\section{Experiments}

In this section, we empirically evaluate our proposed \textbf{Subspace Clifford Networks} across both isotropic and multi-resolution hierarchical architectures. We design our experiments to examine four questions:
\begin{enumerate}
    \item How does the grouped subspace Clifford interaction ($k \in \{2, 3, 4\}$) compare against shift-based rolling CliffordNet \cite{ji2026cliffordnet} in terms of classification accuracy, parameter count, and training time?
    \item What is the effect of varying the subspace group size ($k=2, 3, 4$) and using stage-adaptive hybrid configurations in hierarchical backbones?
    \item How much operator-level speedup and memory reduction does our fused CUDA kernel achieve compared to the native PyTorch implementation?
    \item How does scaling the multi-resolution hierarchical backbone across depth affect accuracy and training cost?
\end{enumerate}

\subsection{Experimental Setup}

\textbf{Dataset, Hardware, and Training Recipe.} Following the evaluation protocol of CliffordNet \cite{ji2026cliffordnet}, we conduct benchmarks on the CIFAR-100 dataset (50,000 training images and 10,000 test images across 100 classes). All models are trained from scratch for 200 epochs using the AdamW optimizer with cosine learning rate decay, a weight decay of $0.1$, and standard data augmentation (AutoAugment and Random Erasing). No pre-trained weights or external data are used. All experiments and operator benchmarks are conducted on a single NVIDIA GeForce RTX 4090 GPU (24\,GB GDDR6X VRAM).

\textbf{Model Variants and Architectural Configurations.} 
We evaluate four model groups in Table~\ref{tab:main_results}, all configured with a total depth of $12$ blocks:
\begin{itemize}
    \item \textbf{Isotropic Models (12 Layers):} The baseline CliffordNet-Nano ($|\mathcal{S}|=2$ shifts) and CliffordNet-Lite ($|\mathcal{S}|=5$ shifts) results are cited from \cite{ji2026cliffordnet}, which use 12-layer isotropic backbones with CUDA-accelerated rolling kernels. Our \textbf{Iso-Subspace-CAN ($k=2$)} and \textbf{Iso-Subspace-CAN ($k=4$)} models use the same 12-layer isotropic macro-topology ($D=128$) with our \texttt{CliffordAlgebraBlock} (Section~\ref{sec:architecture}).
    \item \textbf{Hierarchical Pyramid Models (12 Layers):} Both the \textit{Shift-Based Hierarchical CliffordNet} baselines and our \textbf{Hier-Subspace-CAN} variants share the same 3-stage multi-resolution pyramid macro-architecture, configured with stage depths $(L_1, L_2, L_3) = (3, 4, 5)$ (totaling 12 blocks) and stage channel dimensions $(D_1, D_2, D_3) = (96, 128, 160)$. The shift-based hierarchical baselines use the original GGR block with Sparse Rolling ($\mathcal{S}=\{1, 2, 4\}$ or $\mathcal{S}=\{1, 2, 4, 8\}$), whereas \textbf{Hier-Subspace-CAN} uses our subspace block (Section~\ref{sec:architecture}) with $k \in \{2, 3, 4\}$.
\end{itemize}

\subsection{Main Results and Comparison with Shift-Based CliffordNet}

\begin{table}[th]
\centering
\caption{\textbf{Main Results on CIFAR-100.} Comparison of our Isotropic and Hierarchical Subspace Clifford Networks against shift-based CliffordNet baselines. The isotropic baseline results are cited from \cite{ji2026cliffordnet} (with CUDA-accelerated rolling kernels).}
\label{tab:main_results}
\small
\begin{tabular}{lccccc}
\toprule
\textbf{Model Architecture} & \textbf{Interaction Type} & \textbf{Params} & \textbf{FFN?} & \textbf{Train Time (min)} & \textbf{Top-1 Acc} \\
\midrule
\multicolumn{6}{l}{\textit{CUDA-Accelerated Shift-Based Isotropic CliffordNet Baselines (\texttt{torch.roll}) \cite{ji2026cliffordnet}}} \\
CliffordNet-Nano ($|\mathcal{S}|=2$ shifts) & Sparse Rolling & 1.4M & No & 57.64 & 77.82\% \\
CliffordNet-Lite ($|\mathcal{S}|=5$ shifts) & Sparse Rolling & 2.6M & No & 58.27 & 79.05\% \\
\midrule
\multicolumn{6}{l}{\textit{\color{red!30!blue} Ours: Isotropic Subspace CliffordNet}} \\
\rowcolor{blue!6!white} \textbf{Iso-Subspace-CAN ($k=2$, 12 layers)} & 2D Subspace & \textbf{0.756M} & No & \textbf{29.46} & \textbf{78.32\%} \\
\rowcolor{blue!6!white} \textbf{Iso-Subspace-CAN ($k=4$, 12 layers)} & 4D Subspace & \textbf{0.952M} & No & \textbf{31.54} & \textbf{78.46\%} \\
\midrule
\multicolumn{6}{l}{\textit{Shift-Based Hierarchical CliffordNet (Multi-Resolution Pyramid)}} \\
Hier-CliffordNet ($\mathcal{S}=\{1,2,4\}$) & Sparse Rolling & \textbf{2.093M} & No & \textbf{64.15} & \textbf{79.48\%} \\
Hier-CliffordNet ($\mathcal{S}=\{1,2,4,8\}$) & Sparse Rolling & \textbf{2.535M} & No & \textbf{68.68} & \textbf{79.53\%} \\
\midrule
\multicolumn{6}{l}{\textit{\color{red!30!blue} Ours: Hierarchical Subspace CliffordNet (Multi-Resolution Pyramid)}} \\
\rowcolor{blue!6!white} \textbf{Hier-Subspace-CAN ($k=2$)} & 2D Subspace & \textbf{0.898M} & No & \textbf{41.51} & \textbf{79.48\%} \\
\rowcolor{blue!6!white} \textbf{Hier-Subspace-CAN ($k=3$)} & 3D Subspace & \textbf{1.017M} & No & \textbf{43.24} & \textbf{80.12\%} \\
\rowcolor{blue!6!white} \textbf{Hier-Subspace-CAN ($k=4$)} & 4D Subspace & \textbf{1.120M} & No & \textbf{43.59} & \textbf{80.19\%} \\
\bottomrule
\end{tabular}
\end{table}

Table~\ref{tab:main_results} summarizes the evaluation on CIFAR-100:

\textbf{1. Efficiency in Isotropic Backbones.} 
In the 12-layer isotropic regime, \textbf{Iso-Subspace-CAN ($k=2$)} uses \textbf{0.756M} parameters ($46\%$ fewer than CliffordNet-Nano at $1.4\text{M}$) and achieves a $+0.50\%$ higher Top-1 accuracy ($\mathbf{78.32\%}$ vs.\ $77.82\%$) while reducing training time from $57.64\text{ min}$ to $\mathbf{29.46\text{ min}}$ ($1.96\times$ faster). Increasing the group size to $k=4$, \textbf{Iso-Subspace-CAN ($k=4$)} reaches $\mathbf{78.46\%}$ accuracy with \textbf{0.952M} parameters in $\mathbf{31.54\text{ min}}$, outperforming CliffordNet-Nano by $+0.64\%$. Compared to the larger 5-shift CliffordNet-Lite ($2.6\text{M}$ params, $58.27\text{ min}$, $79.05\%$), Iso-Subspace-CAN ($k=4$) is $0.59\%$ lower in accuracy but uses $63\%$ fewer parameters and trains $1.85\times$ faster.

\textbf{2. Effect of the Multi-Resolution Hierarchical Pyramid.} 
Moving from a single-resolution isotropic column to the 3-stage pyramid with $\texttt{stage\_depths}=(3, 4, 5)$ and $\texttt{stage\_dims}=(96, 128, 160)$ improves accuracy for both interaction types. Under sparse rolling, \texttt{Hier-CliffordNet} ($\mathcal{S}=\{1, 2, 4, 8\}$) reaches $79.53\%$ accuracy, requiring $2.535\text{M}$ parameters and $68.68\text{ min}$ of training time.

\textbf{3. Hierarchical Subspace-CAN Results.} 
Combining our subspace block with the hierarchical pyramid improves the accuracy--cost trade-off across all evaluated group sizes:
\begin{itemize}
    \item At $k=2$, \textbf{Hier-Subspace-CAN ($k=2$)} matches the $\mathbf{79.48\%}$ accuracy of $\texttt{Hier-CliffordNet}$ ($\mathcal{S}=\{1, 2, 4\}$) while reducing parameter count by $\mathbf{57\%}$ ($\mathbf{0.898\text{M}}$ vs.\ $2.093\text{M}$) and training time by $\mathbf{35\%}$ ($\mathbf{41.51\text{ min}}$ vs.\ $64.15\text{ min}$).
    \item At $k=3$, \textbf{Hier-Subspace-CAN ($k=3$)} reaches $\mathbf{80.12\%}$ Top-1 accuracy with $\mathbf{1.017\text{M}}$ parameters—exceeding the $2.535\text{M}$ 4-shift hierarchical baseline by $\mathbf{+0.59\%}$ with \textbf{60\% fewer parameters} and $\mathbf{37\%}$ less training time ($\mathbf{43.24\text{ min}}$ vs.\ $68.68\text{ min}$).
    \item At $k=4$, evaluating all $\binom{4}{2}=6$ within-group wedge pairs yields $\mathbf{80.19\%}$ accuracy ($\mathbf{1.120\text{M}}$ params, $\mathbf{43.59\text{ min}}$), outperforming both isotropic CliffordNet-Lite ($+1.14\%$) and the 4-shift hierarchical baseline ($+0.66\%$).
\end{itemize}

\textbf{4. Depth Scaling Behavior.} 
We further examine scaling total network depth from $12$ to $18$ and $24$ blocks while keeping stage widths fixed at $(D_1, D_2, D_3) = (96, 128, 160)$ and subspace dimensions at $\mathbf{k}=(2, 4, 4)$. As shown in Table~\ref{tab:depth_scaling}, accuracy improves consistently with depth:
\begin{itemize}
    \item The \textbf{18-layer} model ($\texttt{stage\_depths}=(4, 6, 8)$) reaches $\mathbf{80.68\%}$ accuracy with $\mathbf{1.650\text{M}}$ parameters and $57.12\text{ min}$ of training time.
    \item The \textbf{24-layer} model ($\texttt{stage\_depths}=(4, 10, 10)$) reaches $\mathbf{81.21\%}$ accuracy at $\mathbf{2.209\text{M}}$ parameters ($68.97\text{ min}$).
\end{itemize}
Compared to the 12-layer shift-based \texttt{Hier-CliffordNet} ($\mathcal{S}=\{1, 2, 4, 8\}$, $2.535\text{M}$ params, $68.68\text{ min}$, $79.53\%$), our 24-layer \textbf{Hier-Subspace-CAN-24} is twice as deep, uses $13\%$ fewer parameters ($2.209\text{M}$ vs.\ $2.535\text{M}$), trains in similar wall-clock time ($68.97\text{ min}$ vs.\ $68.68\text{ min}$), and improves Top-1 accuracy by $+1.68\%$.

\begin{table}[th]
\centering
\caption{\textbf{Depth Scaling of Hierarchical \texttt{CliffordNet} on CIFAR-100.} All models share the same stage channel widths $\texttt{stage\_dims}=(96, 128, 160)$.}
\label{tab:depth_scaling}
\small
\begin{tabular}{lcccccc}
\toprule
\textbf{Model Variant} & \textbf{Total Depth} & \textbf{Stage Depths} & \textbf{Subspace $\mathbf{k}$} & \textbf{Params (M)} & \textbf{Train Time (min)} & \textbf{Top-1 Acc (\%)} \\
\midrule
Hier-Subspace-CAN-12 & 12 Layers & $(3, 4, 5)$ & $(2, 4, 4)$ & $\mathbf{1.092\text{M}}$ & $42.18$ & $80.25\%$ \\
Hier-Subspace-CAN-18 & 18 Layers & $(4, 6, 8)$ & $(2, 4, 4)$ & $\mathbf{1.650\text{M}}$ & $57.12$ & $80.68\%$ \\
\rowcolor{blue!6!white} \textbf{Hier-Subspace-CAN-24} & \textbf{24 Layers} & $\mathbf{(4, 10, 10)}$ & $\mathbf{(2, 4, 4)}$ & $\mathbf{2.209\text{M}}$ & $\mathbf{68.97}$ & $\mathbf{81.21\%}$ \\
\bottomrule
\end{tabular}
\end{table}

\subsection{Ablation Studies}

To evaluate the effects of subspace group size, hybrid stage schedules, individual branches (inner vs.\ wedge), and the spatial diffusion module, we conduct ablations in Table~\ref{tab:ablation_subspace} using the 12-layer hierarchical backbone ($\texttt{stage\_depths}=(3, 4, 5)$ and $\texttt{stage\_dims}=(96, 128, 160)$).

\begin{table}[th]
\centering
\caption{\textbf{Ablation Studies} on subspace dimension $k \in \{2, 3, 4\}$, stage-adaptive hybrid schedules, interaction branches (inner vs.\ wedge), and spatial diffusion in Hierarchical \texttt{CliffordNet} ($\texttt{stage\_depths}=(3, 4, 5)$, $\texttt{stage\_dims}=(96, 128, 160)$). Bivectors/Group denotes $\binom{k}{2}$; Expansion Ratio denotes $C_{\text{out}}/\tilde{D}$.}
\label{tab:ablation_subspace}
\small
\begin{tabular}{lcccccc}
\toprule
\textbf{Subspace Config} & \textbf{Algebras} & \textbf{Bivectors / Group} & \textbf{Expansion Ratio} & \textbf{Params (M)} & \textbf{Train Time} & \textbf{Top-1 Acc} \\
\midrule
Uniform $k=2$ & $(2, 2, 2)$ & $1$ (2D Area) & $1.5\times$ & \textbf{0.898M} & \textbf{41.51(min)} & \textbf{79.48\%} \\
Uniform $k=3$ & $(3, 3, 3)$ & $3$ (3D Cross-Prod) & $2.0\times$ & \textbf{1.017M} & \textbf{43.24(min)} & \textbf{80.12\%} \\
Uniform $k=4$ & $(4, 4, 4)$ & $6$ (4D Six-Planes) & $2.5\times$ & \textbf{1.120M} & \textbf{43.59(min)} & \textbf{80.19\%} \\
\midrule
Hybrid (234) & $(2, 3, 4)$ & $1 \to 3 \to 6$ & $1.5\times \to 2.0\times \to 2.5\times$ & \textbf{1.061M} & \textbf{42.13(min)} & \textbf{80.22\%} \\
Hybrid (432) & $(4, 3, 2)$ & $6 \to 3 \to 1$ & $2.5\times \to 2.0\times \to 1.5\times$ & \textbf{0.961M} & \textbf{43.62(min)} & \textbf{79.80\%} \\
\rowcolor{green!3!white} Hybrid (244) & $(2, 4, 4)$ & $1 \to 6 \to 6$ & $1.5\times \to 2.5\times \to 2.5\times$ & \textbf{1.092M} & \textbf{42.18(min)} & \textbf{80.25\%} \\
\midrule
Inner only & $(4, 4, 4)$ & $0 \to 0 \to 0$ & $1.0\times$ & \textbf{0.788M} & \textbf{40.39(min)} & \textbf{79.02\%} \\
Wedge only & $(4, 4, 4)$ & $6 \to 6 \to 6$ & $1.5\times$ & \textbf{0.898M} & \textbf{41.20(min)} & \textbf{78.49\%} \\
\midrule
SpatialDiffusion-free & $(4, 4, 4)$ & $6 \to 6 \to 6$ & $2.5\times$ & \textbf{1.075M} & \textbf{31.36(min)} & \textbf{79.02\%} \\
\bottomrule
\end{tabular}
\end{table}

\textbf{1. Uniform vs.\ Stage-Adaptive Hybrid Subspaces.} 
Among uniform configurations, increasing the group size from $k=2$ ($1.5\times$ expansion) to $k=3$ ($2.0\times$) and $k=4$ ($2.5\times$) increases Top-1 accuracy from $79.48\%$ to $80.12\%$ and $80.19\%$. In these experiments, progressive hybrid schedules—\textbf{Hybrid (2, 3, 4)} ($\mathbf{80.22\%}$) and \textbf{Hybrid (2, 4, 4)} ($\mathbf{80.25\%}$)—achieve slightly higher accuracy than uniform $k=4$ ($80.19\%$) while using fewer parameters ($1.061\text{M}$ and $1.092\text{M}$ vs.\ $1.120\text{M}$) and slightly less training time ($42.13\text{ min}$ and $42.18\text{ min}$ vs.\ $43.59\text{ min}$), as using $k_1=2$ in the high-resolution first stage reduces early-stage channel expansion. Conversely, the inverted schedule \textbf{Hybrid (4, 3, 2)} obtains $79.80\%$ accuracy at $43.62\text{ min}$.

\textbf{2. Inner and Wedge Branches.} 
Isolating the two branches at $\mathbf{k}=(4, 4, 4)$ shows that both contribute to final performance. Retaining only the coordinate-wise inner branch (\textbf{Inner only}, $1.0\times$ expansion) or only the anti-symmetric wedge branch (\textbf{Wedge only}, $1.5\times$ expansion) yields $79.02\%$ and $78.49\%$ accuracy, respectively. Combining both branches reaches $\mathbf{80.19\%}$ ($+1.17\%$ over Inner-only and $+1.70\%$ over Wedge-only), confirming that the coordinate-wise products ($\mathbf{u} \odot \mathbf{v}$) and antisymmetric wedge differences ($\mathbf{u} \wedge \mathbf{v}$) provide complementary features.

\textbf{3. Impact of Spatial Diffusion.} 
Removing the \texttt{SpatialDiffusion} module (\textbf{SpatialDiffusion-free}, relying only on the internal $3\times 3$ context depthwise convolution $\operatorname{ctx\_dw}$ for spatial aggregation) achieves $\mathbf{79.02\%}$ accuracy while reducing training time by $\mathbf{28\%}$ (from $43.59\text{ min}$ down to $\mathbf{31.36\text{ min}}$). Including \texttt{SpatialDiffusion} adds $0.045\text{M}$ parameters ($1.075\text{M} \to 1.120\text{M}$) and improves accuracy by $\mathbf{+1.17\%}$ to $80.19\%$.

\subsection{Operator Benchmark: Fused CUDA Kernel vs.\ Native PyTorch}

To quantify the operator-level benefit of our fused CUDA implementation (Section~\ref{sec:cuda_fusion}), we benchmark the standalone \texttt{SubspaceGeometricProduct} module against its native eager-mode PyTorch counterpart (using tensor views, slices, elementwise arithmetic, and `torch.cat`). We measure combined forward + backward execution time and peak extra allocated tensor memory averaged over 100 iterations after warm-up at input shape $(B, D, H, W) = (64, 128, 64, 64)$ in single precision (`float32`), full mode, and differential context ($\lambda=1$).

\textbf{17.2$\times$ Operator Speedup and 50\% Reduction in Peak Extra Allocated Memory.} 
For $k=4$ (where each input tensor $\mathbf{Z}_1, \mathbf{Z}_2$ occupies $128\text{ MiB}$ and the output tensor $\mathbf{F}_{\text{gp}}$ with $320$ channels occupies $320\text{ MiB}$), our fused CUDA kernel reduces the combined forward and backward execution time from $48.67\text{ ms}$ (native PyTorch) to $\mathbf{2.83\text{ ms}}$—a $\mathbf{17.18\times}$ speedup. Under this measurement protocol, peak extra allocated memory drops from $640.0\text{ MiB}$ in native PyTorch to $\mathbf{320.0\text{ MiB}}$ with our fused kernel (matching the $320\text{ MiB}$ allocation of the output tensor itself, as the recomputation-based backward pass avoids allocating global-memory buffers for elementwise intermediates). Numerical checks across all configurations ($k \in \{2, 3, 4\}$, \texttt{full/wedge/inner}, \texttt{diff/abs}) confirm that the maximum absolute difference between our fused CUDA kernel and native PyTorch autograd is within $10^{-7} \sim 10^{-6}$.

\section{Conclusion}

We introduced Subspace Clifford Networks, a reformulation of Clifford algebraic feature interactions that replaces global cyclic channel shifts (\texttt{torch.roll}) with grouped subspace interactions ($k \in \{2, 3, 4\}$) sandwiched between learnable $1\times 1$ linear projections. Within each $k$-dimensional group, the operator evaluates all $\binom{k}{2}$ wedge coordinates together with $k$ $\operatorname{SiLU}$-activated coordinate-wise products, reducing intermediate channel expansion by up to $75\%$ compared to 5-shift rolling. Because the group size $k$ is small and fixed, we implemented a fused CUDA kernel with a recomputation-based backward pass that avoids intermediate elementwise buffers in device memory, achieving a $17.2\times$ operator speedup and a $50\%$ reduction in peak extra allocated memory on the evaluated $k=4$ benchmark. Building on this module, we extended the backbone from single-resolution isotropic columns to multi-resolution Hierarchical Clifford Networks with configurable per-stage subspace dimensions. On CIFAR-100, our 12-layer hierarchical model reaches $80.25\%$ accuracy with $1.09\text{M}$ parameters, and our 24-layer variant reaches $81.21\%$ accuracy while using fewer parameters than 12-layer shift-based baselines.

\bibliographystyle{unsrt} 
\bibliography{sscan}  

\begin{thebibliography}{10}

\bibitem{sandler2018mobilenetv2}
Mark Sandler, Andrew Howard, Menglong Zhu, Andrey Zhmoginov, and Liang-Chieh Chen.
\newblock Mobilenetv2: Inverted residuals and linear bottlenecks.
\newblock In {\em Proceedings of the IEEE/CVF Conference on Computer Vision and Pattern Recognition (CVPR)}, pages 4510--4520, 2018.

\bibitem{ma2018shufflenet}
Ningning Ma, Xiangyu Zhang, Hai-Tao Zheng, and Jian Sun.
\newblock Shufflenet v2: Practical guidelines for efficient {CNN} architecture design.
\newblock In {\em Proceedings of the European Conference on Computer Vision (ECCV)}, pages 116--131, 2018.

\bibitem{dosovitskiy2021image}
Alexey Dosovitskiy, Lucas Beyer, Alexander Kolesnikov, Dirk Weissenborn, Xiaohua Zhai, Thomas Unterthiner, Mostafa Dehghani, Matthias Minderer, Georg Heigold, Sylvain Gelly, Jakob Uszkoreit, and Neil Houlsby.
\newblock An image is worth 16x16 words: Transformers for image recognition at scale.
\newblock In {\em International Conference on Learning Representations (ICLR)}, 2021.

\bibitem{touvron2021training}
Hugo Touvron, Matthieu Cord, Matthijs Douze, Francisco Massa, Alexandre Sablayrolles, and Herv{\'e} J{\'e}gou.
\newblock Training data-efficient image transformers \& distillation through attention.
\newblock In {\em International Conference on Machine Learning (ICML)}, pages 10347--10357, 2021.

\bibitem{gu2023mamba}
Albert Gu and Tri Dao.
\newblock Mamba: Linear-time sequence modeling with selective state spaces.
\newblock {\em arXiv preprint arXiv:2312.00752}, 2023.

\bibitem{zhu2024vision}
Lianghui Zhu, Bencheng Liao, Qian Zhang, Xinlong Wang, Wenyu Liu, and Xinggang Wang.
\newblock Vision mamba: Efficient visual representation learning with bidirectional state space model.
\newblock In {\em Proceedings of the 41st International Conference on Machine Learning (ICML)}, 2024.

\bibitem{liu2024vmamba}
Yue Liu, Yunjie Tian, Yuzhong Zhao, Hongtian Yu, Lingxi Xie, Yaowei Wang, Qixiang Ye, Jianbin Jiao, and Yunfan Liu.
\newblock Vmamba: Visual state space model.
\newblock In {\em Advances in Neural Information Processing Systems (NeurIPS)}, volume~37, 2024.

\bibitem{yu2022metaformer}
Weihao Yu, Mi~Luo, Pan Zhou, Chenyang Si, Yichen Zhou, Xinchao Wang, Jiashi Feng, and Shuicheng Yan.
\newblock Metaformer is actually what you need for vision.
\newblock In {\em Proceedings of the IEEE/CVF Conference on Computer Vision and Pattern Recognition (CVPR)}, pages 10819--10829, 2022.

\bibitem{ji2026cliffordnet}
Zhongping Ji.
\newblock Cliffordnet: All you need is geometric algebra.
\newblock {\em arXiv preprint arXiv:2601.06793}, 2026.

\bibitem{hestenes1966space}
David Hestenes.
\newblock {\em Space-Time Algebra}.
\newblock Gordon and Breach, 1966.

\bibitem{dorst2007geometric}
Leo Dorst, Daniel Fontijne, and Stephen Mann.
\newblock {\em Geometric Algebra for Computer Science: An Object-Oriented Approach to Geometry}.
\newblock Morgan Kaufmann, 2007.

\bibitem{brandstetter2023clifford}
Johannes Brandstetter, Rianne van~den Berg, Max Welling, and Jayesh~K. Gupta.
\newblock Clifford neural layers for {PDE} modeling.
\newblock In {\em International Conference on Learning Representations (ICLR)}, 2023.

\bibitem{brehmer2023geometric}
Johann Brehmer, Pim de~Haan, S{\"o}nke Behrends, and Taco Cohen.
\newblock Geometric algebra transformer.
\newblock In {\em Advances in Neural Information Processing Systems (NeurIPS)}, volume~36, pages 35472--35496, 2023.

\bibitem{liu2022convnet}
Zhuang Liu, Hanzi Mao, Chao-Yuan Wu, Christoph Feichtenhofer, Trevor Darrell, and Saining Xie.
\newblock A convnet for the 2020s.
\newblock In {\em Proceedings of the IEEE/CVF Conference on Computer Vision and Pattern Recognition (CVPR)}, pages 11976--11986, 2022.

\bibitem{dao2022flashattention}
Tri Dao, Daniel~Y. Fu, Stefano Ermon, Atri Rudra, and Christopher R{\'e}.
\newblock Flashattention: Fast and memory-efficient exact attention with io-awareness.
\newblock In {\em Advances in Neural Information Processing Systems (NeurIPS)}, volume~35, pages 16344--16359, 2022.

\end{thebibliography}

\end{document}